\documentclass[10pt,twocolumn,letterpaper]{article}

\usepackage{iccv}
\usepackage{times}
\usepackage{epsfig}
\usepackage{graphicx}
\usepackage{amsmath}
\usepackage{amssymb} % amsfonts

\usepackage{amsthm}
\usepackage{cite}
\usepackage{multirow}
\usepackage{commath}
\usepackage[switch]{lineno}
\usepackage{enumitem}

\DeclareMathOperator*{\argmax}{argmax}
\DeclareMathOperator*{\argmin}{argmin}
\newcommand{\R}{\mathbb{R}}

\DeclareMathOperator*{\KL}{KL}
\DeclareMathOperator*{\CE}{CE}

\newtheorem{theorem}{Theorem}[section]

\usepackage[pagebackref=true,breaklinks=true,letterpaper=true,colorlinks,bookmarks=false]{hyperref}

\iccvfinalcopy % *** Uncomment this line for the final submission

\def\iccvPaperID{-1} % *** Enter the ICCV Paper ID here

\ificcvfinal\fi

\begin{document}

    %%%%%%%%% TITLE
    \title{ARM: A Confidence-Based Adversarial Reweighting Module for Coarse Semantic Segmentation}

    \author{
        Jingchao Liu, Ye Du, Zehua Fu, Qingjie Liu, Yunhong Wang\\
        \texttt{\small{\{liu.siqi, duyee, qingjie.liu, yhwang\}@buaa.edu.cn}, zehua\_fu@163.com}
    }

    \maketitle
    \ificcvfinal\thispagestyle{empty}\fi

    %%%%%%%%% ABSTRACT
    \begin{abstract}
        Coarsely-labeled semantic segmentation annotations are easy to obtain, but therefore bear the risk of losing edge details and introducing background pixels.
        Impeded by the inherent noise, existing coarse annotations are only taken as a bonus for model pre-training.
        In this paper, we try to exploit their potentials with a confidence-based reweighting strategy.
        To expand, loss-based reweighting strategies usually take the high loss value to identify two completely different types of pixels, namely, valuable pixels in noise-free annotations and mislabeled pixels in noisy annotations.
        This makes it impossible to perform two tasks of mining valuable pixels and suppressing mislabeled pixels at the same time.
        However, with the help of the prediction confidence, we successfully solve this dilemma and simultaneously perform two subtasks with a single reweighting strategy.
        Furthermore, we generalize this strategy into an Adversarial Reweighting Module (ARM) and prove its convergence strictly.
        Experiments on standard datasets shows our ARM can bring consistent improvements for both coarse annotations and fine annotations.
        Specifically, built on top of DeepLabv3+, ARM improves the mIoU on the coarsely-labeled Cityscapes by a considerable margin and increases the mIoU on the ADE20K dataset to \textbf{47.50}.
    \end{abstract}

    \section{Introduction}\label{sec:introduction}
\begin{figure}
    \centering
    \includegraphics[width=0.9\columnwidth]{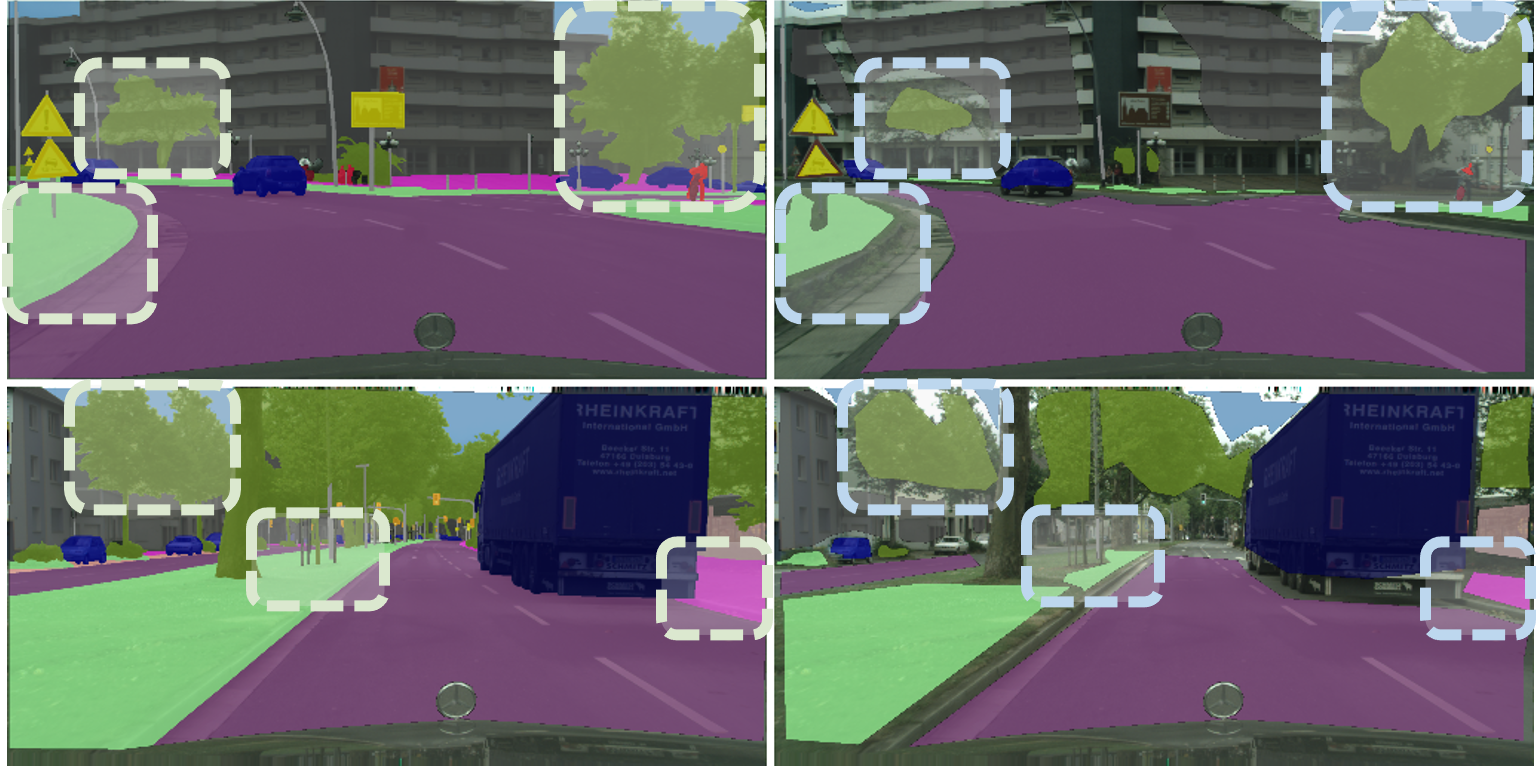}
    \caption{Fine annotation \textit{vs} coarse annotation.
    Different from the existing fine annotation, the coarse annotation only labels objects with coarse polygons.
    To show the difference more clearly, several regions have been marked.
    }
    \label{fig:fine-vs-coarse}
\end{figure}

    Driven by tons of high-quality dense-annotated images, existing segmentation models (e.g.~\cite{chen2018encoder,takikawa2019gated,yuan2019object}) have achieved remarkable performance.
However, the labeling cost is still unaffordable for practical applications that rely on large-scale data.
As reported by Cityscapes\cite{cordts2016cityscapes}, the labeling time for only 5,000 images has reached 7,500 hours.
To provide more annotations with less cost, Cityscapes introduces the \textit{coarse annotation}.
As shown in Figure~\ref{fig:fine-vs-coarse}, different from the existing fine annotation which aims to label each pixel accurately, the coarse annotation only labels objects with coarse polygons.
By using it, Cityscapes shortens the labeling time of a single image from 1.5 hours to 7 minutes and provides up to 20,000 images with only 2,333 hours.

Although these coarsely-labeled images are easy to obtain, the inherent label noise makes them challenging to be exploited.
As a workaround, existing methods usually only take coarse annotations as a bonus for model pre-training and then perform fine-tuning on fine annotations to obtain the optimal results.
Inspired by the recent progress in segmentation with box-level annotations~\cite{khoreva2017simple,kulharia2020box2seg,tian2020boxinst} and the analysis of the coarse annotation in~\cite{zlateski2018importance,luo2018coarse}, we want to examine whether coarse annotations have the potential to generate high-quality semantic segmentations, which may help us to escape from the expensive annotation cost and further benefits to practical applications.
Working towards this goal, we define the task \textit{Coarse Semantic Segmentation} as training a model only with coarsely-labeled images.

\begin{figure}
    \centering
    \includegraphics[width=0.7\columnwidth]{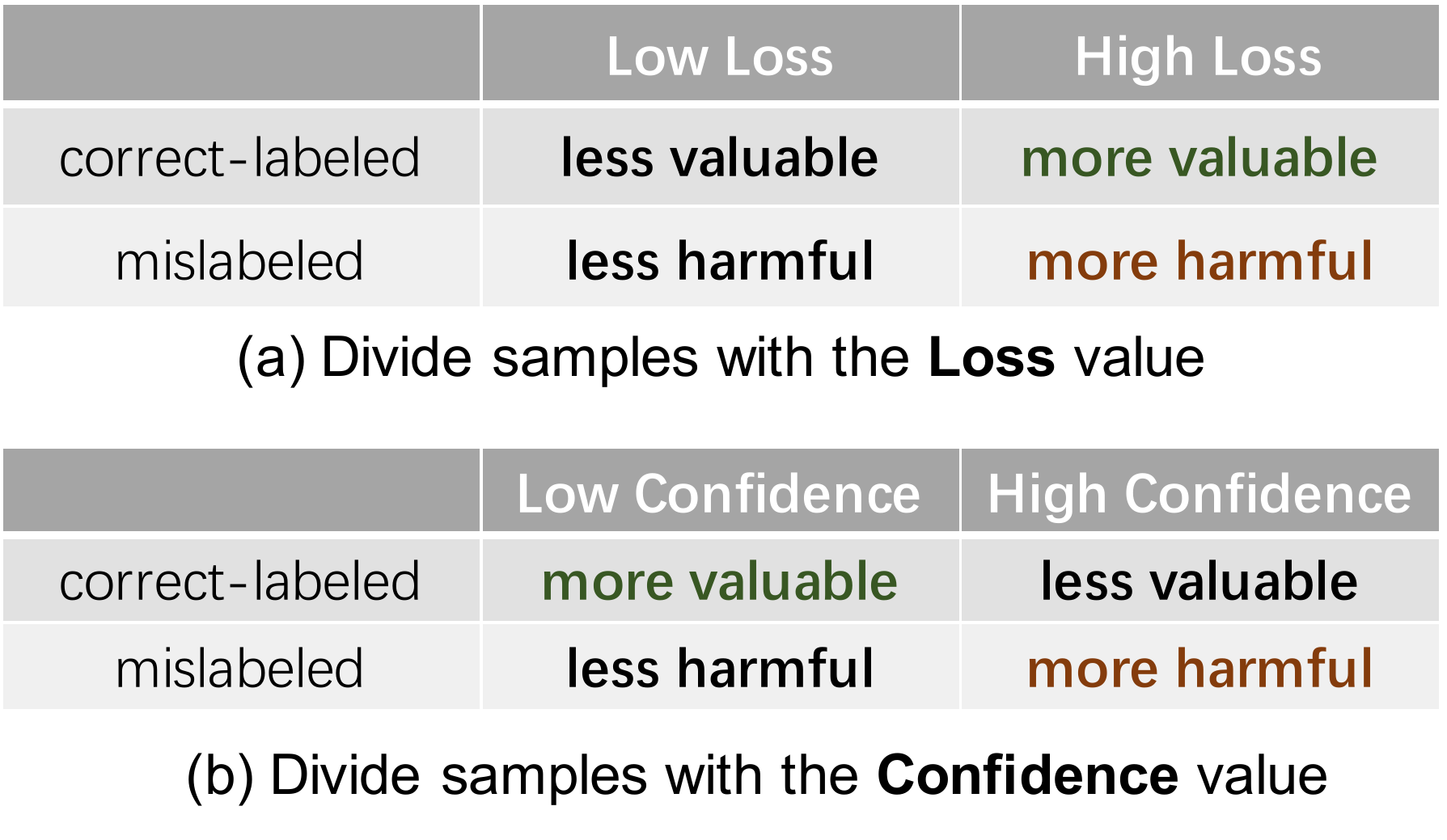}
    \caption{
        Divide samples with different indicators.
        The confidence value can distinguish more valuable samples from more harmful samples while the loss value cannot.
    }
    \label{fig:reweight-by-confidence}
\end{figure}

In this paper, we try to treat each image pixel as an independent sample and then suppress the pixel-wise label noise with a reweighting strategy.
However, the following observations put us into a dilemma.
In noise-free datasets, methods like OHEM~\cite{shrivastava2016training} and Focal loss~\cite{lin2017focal} usually identify valuable hard samples by the high loss value.
while in noisy datasets, methods like GCE~\cite{zhang2018generalized} also identify harmful mislabeled samples by the high loss value.
After dividing samples according to the loss value and the correctness of label, we can put samples into the grids of Figure~\ref{fig:reweight-by-confidence}(a).
As we can see, high-loss samples actually contains two types of samples at the same time, namely, correctly-labeled valuable samples and mislabeled harmful samples.
While focusing on high-loss samples will suffer from noisy labels, an opposite operation will ignore many valuable samples.

To solve this dilemma, we propose to divide samples by the \textit{prediction confidence}, namely, how confident the model thinks which category a sample should belong to.
We have the following observations:
For noise-free datasets, high loss means hard-to-classify, in other words, valuable samples identified by the high loss value actually are those low-confidence ones.
For noisy datasets, high loss means easy-to-classify but mislabeled, in other words, harmful samples identified by the high loss value actually are those high-confidence ones.
For a better understanding, we re-divide samples according to the confidence value in Figure~\ref{fig:reweight-by-confidence}(b).
As we can see, different from Figure~\ref{fig:reweight-by-confidence}(a), valuable samples and harmful samples now can be distinguished by the confidence value.
Furthermore, we find high-confidence samples are either less valuable or more harmful.
For correctly-labeled samples, high confidence means the model can easily classify the sample to the correct category.
Since mining hard samples usually means suppressing easy samples, with the same goal of OHEM, we think these high-confidence samples are less valuable and should be suppressed.
On the other hand, once these high-confidence samples are mislabeled, a large classification loss will seriously interfere with the training process.
Thus, suppressing high-confidence samples can not only act as OHEM in both noise-free and noisy datasets, but also can simultaneously suppress mislabeled samples in noisy datasets.

Unfortunately, our thesis shows the existing confidence indicator introduced by KL loss~\cite{kendall2017uncertainties} cannot produce robust results as expected.
In this work, we resort to the pixel-wise class-likelihoods $p_{1:C}$ and take its variance as an acceptable confidence indicator.
Note that, our work's contribution is mainly on the reweighting strategy ``suppressing high-confidence samples'' instead of designing an optimal confidence indicator.
Any existing confidence indicator can be applied to our reweighting strategy.

Our second contribution is oriented with how to map the confidence into a sample weight properly.
Considering existing weight mapping functions usually require extensive experiments for searching the optimal hyper-parameters, we propose a reweighting module named Adversarial Reweighting Module (ARM) and strictly prove its convergence.
Our ARM has the following advantages:

\begin{itemize}[noitemsep]
    \item Adopting the idea of suppressing high-confidence samples, ARM is a robust module and can be applied to both noisy datasets and noise-free datasets.
    \item Guaranteed by the proof of convergence, ARM can automatically derive the optimal weight mapping function with only a single experiment.
    \item Designed with 53 parameters, ARM is lightweight and only brings negligible computational overhead.
    \item Performing operation on the final class-likelihood map, ARM is model-independent and can be applied any pixel-wise loss function.
    \item Just changing the learning rate of ARM to a negative value is enough to perform the adversarial training.
\end{itemize}

A similar work to ours is the MetaWeight~\cite{shu2019meta}, yet, our ARM is fundamentally different from theirs.
While their reweighting module aims to \textbf{minimize} the loss on an extra \textbf{validation dataset}, our ARM manages to \textbf{maximize} the loss on the \textbf{training dataset}.
See more details about our ARM in Sec.~\ref{subsec:adversarial-reweighting-module} and Sec.~\ref{subsec:convergence-of-ARM}.
Extensive experiments are conducted on two semantic segmentation benchmarks, namely, Cityscapes~\cite{cordts2016cityscapes} and ADE20k~\cite{zhou2017scene}.
Results on both coarse annotations and fine annotations demonstrate the robustness and effectiveness of our ARM.
Specifically, built on top of DeepLabv3+, we improve the mIoU on the coarsely-labeled Cityscapes by a considerable margin and increase the mIoU on the ADE20K dataset to \textbf{47.50}.

    \section{Related Work}\label{sec:related-work}
    \textbf{Label Noise.}
Dealing with label noise is a long-standing problem and is mainly studied in the image classification field.
A typical solution is identifying noisy samples and then correct these wrong labels~\cite{li2017learning}, or directly assigning smaller weights to them to reduce noise impact~\cite{ren2018learning,wang2017robust}.
An alternative strategy involving training samples attempt to reorganize sampling priority and frequency of samples~\cite{chang2017active,jiang2018mentornet}.
Another class of methods tries to design robust loss functions inherently tolerant to noisy labels~\cite{zhang2018generalized,ma2020normalized}.
By assuming the noisy labels are corrupted good ones by a noise transition matrix and embedding the estimated matrix into loss functions,~\cite{han2018masking} also relieves noise burden.
A detailed overview on this topic can be found in~\cite{song2020learning}.

\textbf{Reweighting Strategy.}
Reweighting strategies play an important role in improving model performance.
The loss value is easily accessible and can be taken as a good indicator used in reweighting strategies.
However, there exist two entirely contradicting ideas when talking about reweighting by loss.
To mine valuable hard samples in noise-free datasets, OHEM~\cite{shrivastava2016training} and Focal loss~\cite{lin2017focal} assign samples with high loss a large weight;
while, to suppress mislabeled samples in noisy datasets, robust loss functions like~GCE\cite{zhang2018generalized} tend to assign samples with high loss a small weight or gradient.
As can be seen, reweighting by loss cannot distinguish hard samples from mislabeled samples.
To solve this issue, researchers in the meta-learning field try to introduce a noise-free validation set and then reweight according to whether the sample can increase the performance on this validation set.
~\cite{shu2019meta} and~\cite{ren2018learning} determine the sample weight by the loss value and the gradient value on the validation set, respectively.
Apart from these methods,~\cite{chang2017active} assigns weight to a sample based on its variance of predicted probabilities collected from previous epochs.
In the object detection task, sample reweighting is also an effective method, e.g., PISA~\cite{cao2020prime} proposes to reweight anchors by IoU and develop a sampling strategy called Prime Sample Attention.

\textbf{KL Loss.}
Kendall et al.~\cite{kendall2017uncertainties} make an in-depth analysis on KL Divergence and then propose the KL loss to explicitly modeling prediction uncertainty for the regression task.
Later, they show that this idea is also applicable to classification tasks~\cite{kendall2018multi}.
Following this spirit, many works study KL loss and intend to leverage it for boosting performance.
For example, in the object detection field,~\cite{he2019bounding} introduces KL loss to simultaneously learn the bounding box offset and the localization uncertainty and further performs a soft NMS based on this predicted uncertainty.
Taking a step further,~\cite{cai2020learning} jointly learns sample weights for both classification and regression tasks.
Recently,~\cite{zheng2020rectifying} introduces KL loss into cross-domain semantic segmentation.
They explicitly estimate the prediction uncertainty and then use it as a cue to rectify the pseudo-label learning.

    \section{Methodology}\label{sec:methodology}
\begin{figure}
    \centering
    \includegraphics[width=0.95\columnwidth]{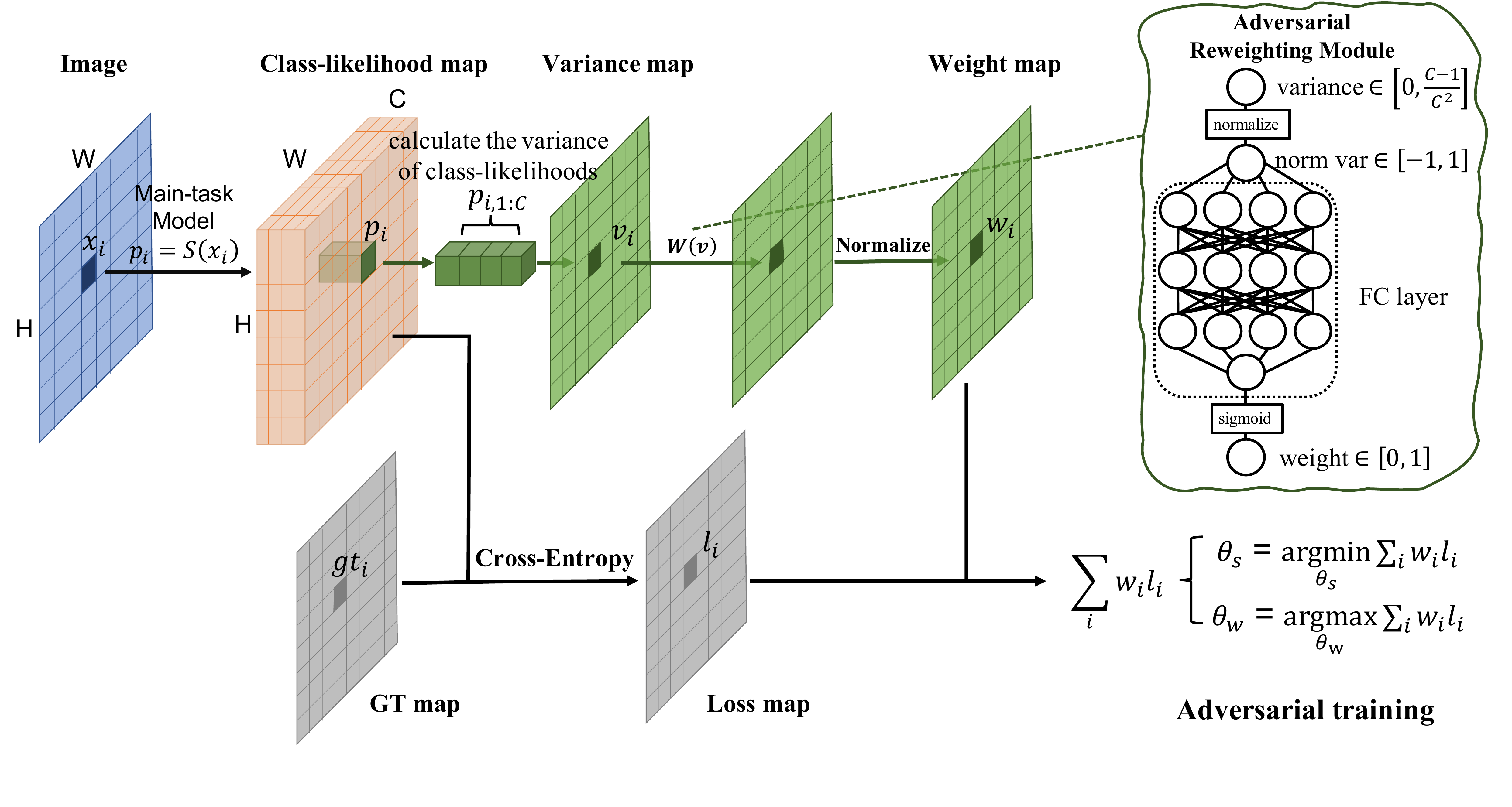}
    \caption{The training pipeline after applying our ARM. After the main-task (segmentation) model generates a \textit{class-likelihood map} for each image, we manually calculate the variance of the predicted class-likelihoods in pixel-wise and then forward it to our ARM to produce a \textit{weight map}. By applying the normalized \textit{weight map} to \textit{loss map}, we get a weighted loss sum. While the main-task model aims to minimize this weighted sum by editing the \textit{class-likelihood map}, our ARM aims to maximize the weighted sum by editing the \textit{weight map}.}
    \label{fig:pipeline}
\end{figure}

    In this section, we start with a brief analysis of the weakness of KL loss and then propose another acceptable confidence indicator \textit{variance}.
After that, we generalize the strategy ``suppressing high-confidence samples'' to an adversarial reweighting module named ARM and strictly prove its convergence.
Finally, we give an example pipeline and illustrate how to use our ARM in segmentation task.
It is recommended to immediately reading Figure~\ref{fig:pipeline} to form a general understanding before continuing.

\subsection{Dive into KL Loss}\label{subsec:dive-into-kl-loss}
KL loss for classification task is defined in~\cite{kendall2018multi} as
\begin{linenomath}
    \begin{gather}
        \KL(p_{1:C}, s, gt) = \frac{1}{s}\CE(p_{1:C}, gt)+\frac{1}{2}\log s,\label{eq:kl-loss-cls}.
    \end{gather}
\end{linenomath}
where $\CE(p_{1:C}, gt)$ indicates the cross-entropy (CE) loss between class-likelihoods $p_{1:C}$ and ground-truth label $gt$, and $s$ indicates the predicted uncertainty.
By minimizing KL loss with regard to $s$ in Eq.~\ref{eq:kl-with-regrad-to-s}, we can find the optimal $\hat{s}$ is related to the likelihood in the ground-truth channel $p_{gt}$.
\begin{equation}
    \begin{gathered}
        \pd{\KL(p_{1:C}, s, gt)}{s} = \frac{s - 2\CE(p_{1:C}, gt)}{2s^2}, \\
        \CE(p_{1:C}, gt) = -\log p_{gt}, \\
        \Rightarrow \hat{s} = \argmin_{s \in {\R}^+} \KL(p_{1:C}, s, gt)= -2\log p_{gt}.
    \end{gathered} \label{eq:kl-with-regrad-to-s}
\end{equation}
A well-trained model on a noise-free dataset tends to act as:
$p_{gt} \rightarrow 1 \Rightarrow s \rightarrow 0\ \ and\ \ p_{gt} \rightarrow 0 \Rightarrow s \rightarrow +\infty$.
Since $1/s$ in Eq.~\ref{eq:kl-loss-cls} is a loss weight of CE loss, a high $p_{gt}$ actually leads to a high sample weight $1/s$.
In other words, contrary to the reweighting strategy we advocate, \textbf{KL loss will assign larger weights to high-confidence samples}.
This reweighting strategy may work well on noise-free datasets, however, applying it to noisy datasets tend to be very fragile, since there may exist an unexpected inconsistency between the expected $\widehat{gt}$ and the labeled $\widetilde{gt}$.
Once a high-confidence sample is mislabeled, the low labeled $\widetilde{p}_{gt}$ will produce a large loss $-\log \widetilde{p}_{gt}$ while the high expected $\widehat{p}_{gt}$ will further assign this sample a big weight.
Therefore, KL loss tends to enlarge the abnormal gradient of a mislabeled high-confidence sample and is not robust to noisy datasets.

\subsection{Variance as a Confidence Indicator}\label{subsec:variance-as-a-confidence-indicator}
Alternatively, we take the variance of the pixel-wise class-likelihoods $p_{1:C}$ as a confidence indicator.
We have
\begin{linenomath}
    \begin{equation}
        \begin{gathered}
            \sum_{c} p_c = 1, p_c \in \left[ 0,1 \right], \bar{p} = \frac{1}{C}\\
            var = \sum_{c} (p_c - \bar{p})^2 \in \left[ 0, \frac{C-1}{C^2} \right].
        \end{gathered} \label{eq:variance-definition}
    \end{equation}
\end{linenomath}

Two inspiring observations can be made from Eq.~\ref{eq:variance-definition}:
First, $var_{\min}$ is obtained when equal likelihoods are predicted over all classes, which means the model does not know which class the pixel belongs to.
Second, $var_{\max}$ is obtained when the likelihood of a specific class is predicted to $1$, which means the model has absolute confidence.
Consequently, $var$ is positively related to predicted confidence.

\begin{figure}
    \centering
    \includegraphics[width=0.9\columnwidth]{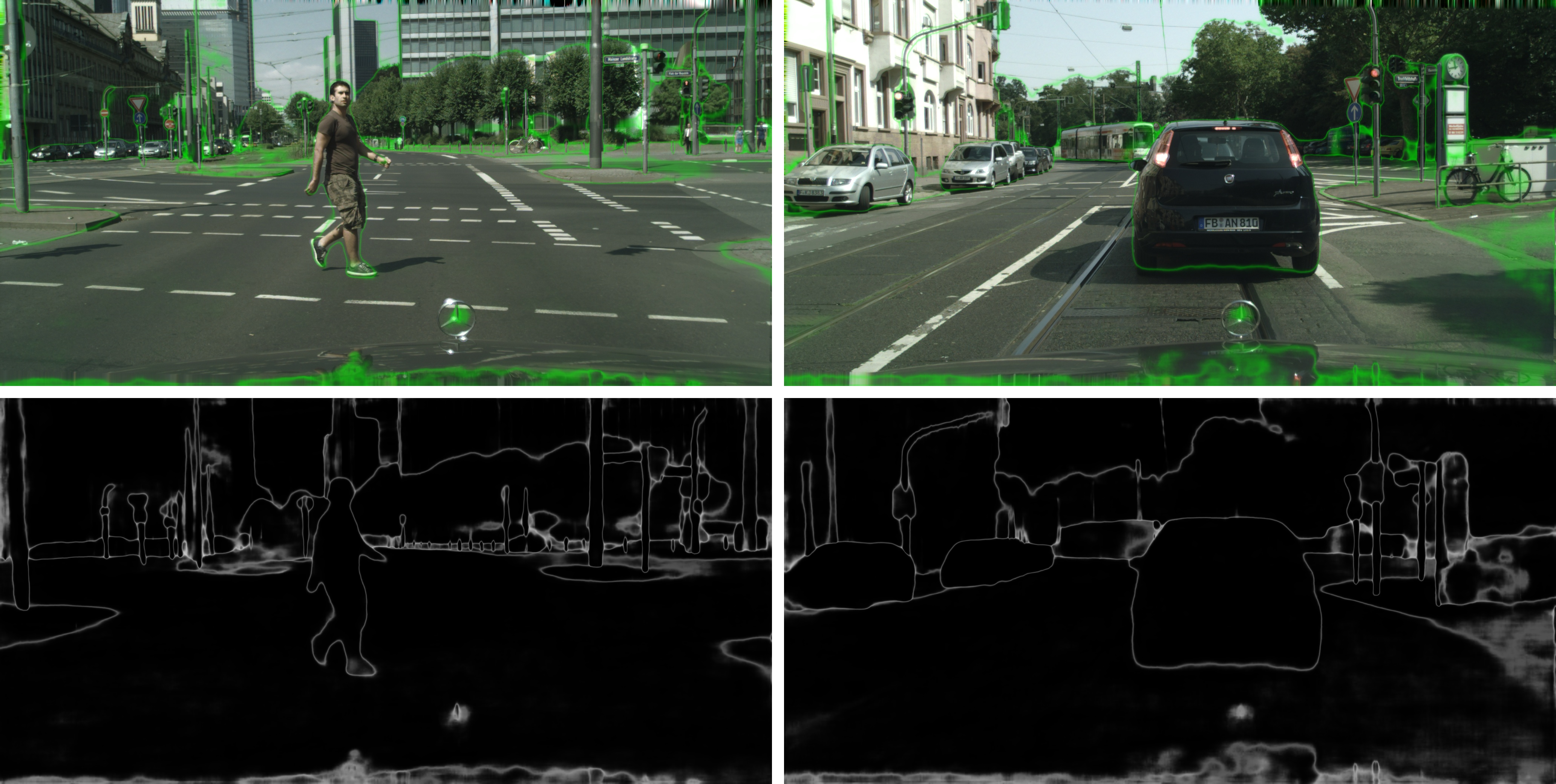}
    \caption{Illustration of low-variance pixels in real images.
    First row: low-variance pixels are painted with green color in real images.
    Second row: low-variance pixels are assigned greater brightness in intensity maps.}
    \label{fig:variance-map}
\end{figure}

We also visualize the low-variance region with images in the Cityscapes dataset.
The segmentation model we use is only supervised with the cross-entropy loss without any tricks.
As shown in Figure~\ref{fig:variance-map}, almost all low-variance pixels locate in the boundaries of adjacent semantic regions,
This is consistent with the consensus that segmentation models are weak at generating clear boundaries and also proves the variance is a good confidence indicator.

Since high-confidence samples can be identified by their variance values, it is natural to transform the training strategy ``suppressing high-confidence samples'' to ``suppressing high-variance samples''.
Moreover, this transformation makes it possible to give a solid proof for the rationality of ``suppressing high-variance samples'' (See our Appendix).

Note that, the definition of our variance is different from all methods in Sec.~\ref{sec:related-work}.
The variance in~\cite{kendall2017uncertainties,kendall2018multi,he2019bounding,cai2020learning} is obtained from the $s$ term in KL loss.
The variance in~\cite{zheng2020rectifying} is calculated among two feature maps outputted from different heads.
The variance in~\cite{chang2017active} is calculated from its historical probability values in previous epochs.
However, our variance is derived from a single feature map and is calculated among the pixel-wise class-likelihoods.

\subsection{Adversarial Reweighting Module}\label{subsec:adversarial-reweighting-module}
To illustrate the proposed reweighting strategy concretely, we formulate the training process as follows
\begin{linenomath}
    \begin{equation}
        \begin{aligned}
            p_{i,1:C} &= S(x_i; \theta_S), \\
            w_i &= W(v_i; \theta_W), \\
            l_i &= L(p_{i,1:C}, gt_i), \\
            \widehat{\theta}_S &= \argmin_{\theta_s} \sum_{i} w_{i}l_{i}.
        \end{aligned} \label{eq:formulate-strategy}
    \end{equation}
\end{linenomath}
Firstly, a segmentation model $S(x)$ controlled by parameters $\theta_S$ predicts the class-likelihoods $p_{i,1:C}$ for each pixel $x_i$.
Then a weight mapping function $W(v)$ controlled by parameters $\theta_W$ generates the weight $w_i$ from the pixel-wise variance $v_i$.
After that, the pixel-wise loss $l_i$ is calculated from the $p_{i,1:C}$ and the $gt_i$.
Finally, the optimal $\theta_S$ is obtained by minimizing the weighted sum of the $w_i$ and $l_i$.

Considering all samples with $v_i = v$ share a same weight $W(v)$, after gathering all samples with $v_i = v$, we define the sum of their loss values as
\begin{linenomath}
    \begin{gather}
        L(v) = \sum_{\{i|v_i=v\}} l_i.\label{eq:l-v}
    \end{gather}
\end{linenomath}
By doing this, the optimization target can be transformed to
\begin{linenomath}
    \begin{equation}
        \begin{gathered}
            \sum_{\{i|v_i=v\}} w_{i}l_{i} = W(v)L(v) \\
            Q = \sum_{i} w_{i}l_{i} = \int W(v;\theta_W)L(v;\theta_S) \text{d}v.\label{eq:formulate-strategy-2}
        \end{gathered}
    \end{equation}
\end{linenomath}

Having laid the groundwork in Eq.~(\ref{eq:formulate-strategy}-\ref{eq:formulate-strategy-2}), we will now analyze $W(v;\theta_W)$ and $L(v;\theta_S)$ separately.

$W(v;\theta_W)$ is determined by our reweighting strategy.
The strategy ``suppressing high-variance samples'' requires $W(v)$ to assign samples with larger $v$ a smaller $w_i$, namely,
\begin{linenomath}
    \begin{gather}
        v_1 > v_2 \implies W(v_1) < W(v_2).\label{eq:reweighting-strategy}
    \end{gather}
\end{linenomath}

\begin{figure}
    \centering
    \includegraphics[width=\columnwidth]{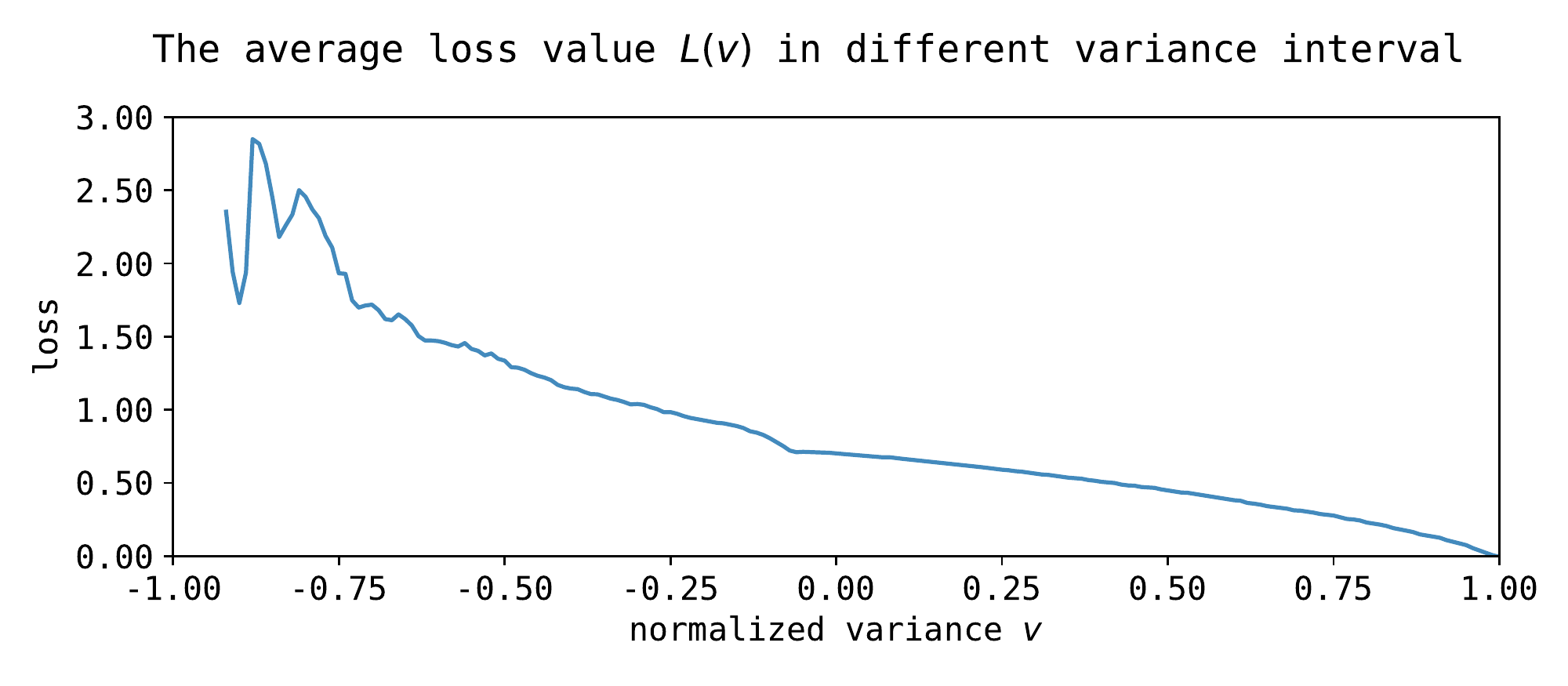}
    \caption{The average loss value in different variance intervals.
    The variance $v$ is normalized to $\frac{2C^2}{C-1}v - 1 \in [-1, 1]$.}
    \label{fig:loss-value-condition-on-var-histogram}
\end{figure}

$L(v;\theta_S)$ is determined by our segmentation model and indicates the sum of loss values in different variance intervals.
To figure out the curve shape of $L(v;\theta_S)$, we make a statistic on the Cityscapes dataset.
Since we usually do iterative training on the mini-batch, the statistic is made on the expectation (average) value of loss values, instead of the sum of them.
As shown in Figure~\ref{fig:loss-value-condition-on-var-histogram}, $L(v;\theta_S)$ acts as
\begin{linenomath}
    \begin{gather}
        v_1 > v_2 \implies L(v_1) < L(v_2).\label{eq:loss-distribution}
    \end{gather}
\end{linenomath}
which means the average loss of low-confidence intervals is bigger than those in high-confidence intervals.

Based on Eq.~(\ref{eq:reweighting-strategy}-\ref{eq:loss-distribution}), we can say $W(v; \theta_W)$ actually tends to focus on the high $L(v)$ interval.
In more detail, while $S(x; \theta_S)$ targets on predicting more accurate $p_{i,1:C}$ and producing lower weighted sum $Q$;
$W(v; \theta_W)$ tends to assign larger weights to samples with lower confidence and produce higher $Q$.
Thus, we introduce the adversarial training strategy as
\begin{linenomath}
    \begin{equation}
        \begin{gathered}
            \theta_s = \argmin_{\theta_s} Q,\\
            \theta_w = \argmax_{\theta_w} Q.\\
        \end{gathered}\label{eq:optimize-SW}
    \end{equation}
\end{linenomath}
and name $W(v; \theta_W)$ as our ARM.

Although Eq.~\ref{eq:optimize-SW} seems difficult to solve, a vanilla SGD is actually sufficient.
There are two key points to simplify this problem.
First, $\theta_S$ can only adjust the $L(v)$, and $\theta_W$ can only adjust the $W(v)$.
$\theta_S$ and $\theta_W$ is independent and parallel.
Second, for continuous functions, the fastest increasing direction is the opposite of the fastest decreasing direction.
While assigning a positive learning rate to $\theta_S$ tends to minimize $Q$, assigning a negative learning rate to $\theta_W$ tends to maximize $Q$.
Therefore, just changing the learning rate of $\theta_W$ to a negative value and minimizing $Q = \sum_{i} w_{i}l_{i}$ as usual is enough to solve Eq.~\ref{eq:optimize-SW} and implements our ARM.

\textbf{What is ARM?}
Our ARM is a learnable mapping function $W(v; \theta_W)$.
Its input is a pixel-wise normalized variance $v_i \in [-1, 1]$.
Its output is a pixel-wise sample weight $w_i \in [0, 1]$.
Its curve shape is controlled by the parameter $\theta_W$.
Existing weight mapping functions are usually designed to a fixed representations like $(av + b)$ or $(a\log v + b)$ and then grid-search the hyper-parameter $a$ and $b$ with \textbf{multiple} experiments.
However, our ARM can embed the mapping function into a multilayer perceptron, and obtain the optimal $W(v; \theta_W)$ with only a \textbf{single} experiment.

\subsection{Convergence of ARM}\label{subsec:convergence-of-ARM}
We can start with an easy but fundamental problem.
Let $l_1=1.2, l_2=0.6, l_3=0.1$, if we constrain $\norm{W}_p = (w_1^p + w_2^p + w_3^p)^{\frac{1}{p}} = 1$ with $p \geq 1$, what is the solution of
\begin{linenomath}
    \begin{gather}
        w_1, w_2, w_3 = \argmax_{\norm{W}_p = 1, w_i > 0} w_{1}l_{1}+w_{2}l_{2}+w_{3}l_{3}. \label{eq:lp-question}
    \end{gather}
\end{linenomath}
We can find that $p=1$ produces a sparse solution
\begin{linenomath}
    \begin{gather}
        w_1: w_2: w_3 = 1: 0: 0, \label{eq:solution-l1}
    \end{gather}
\end{linenomath}
which indicates $l_1$ obtains the total weight;
while $p>1$ produces a smooth solution
\begin{linenomath}
    \begin{gather}
        w_1: w_2: w_3 = 1.2^\frac{1}{p-1}: 0.6^\frac{1}{p-1} : 0.1^\frac{1}{p-1}, \label{eq:solution-lp}
    \end{gather}
\end{linenomath}
which indicates all of $l_1$, $l_2$ and $l_3$ will get a weight.

The above analysis gives us an intuition that different constraints on $W$ will lead to different solutions, and applying L1 Norm ($p=1$) to $W$ will cause the model only focus on the hardest sample, which is very fragile.
Thus, it is necessary to constrain $\norm{W(v;\theta_W)}_p$ with $p > 1$.
In fact, the theorem ``Holder's inequality''\cite{holder1889ueber} provides a detailed analysis for optimization problem like Eq.~\ref{eq:arm-converge}.
The definition of $W(v)$ and $L(v)$ are consistent with the previous section.
\begin{linenomath}
    \begin{gather}
        W(v) = \argmax_{\norm{W}_p=1, p > 1} \int W(v)L(v) \text{d}v,\label{eq:arm-converge}
    \end{gather}
\end{linenomath}

\begin{theorem}[Holder's inequality]
    ~\\
    Let $(S, \Sigma, \mu)$ be a measure space and let $p, q \in (1, \infty)$ with $\frac{1}{p} + \frac{1}{q} = 1$.
    For all measurable functions $f$ and $g$, we have
    \[
        \norm{fg}_1 \leq \norm{f}_p\norm{g}_q
    \]
    where $\norm{f}_p = \left(\int_{\R} \abs{f(x)}^p \text{d}x\right)^\frac{1}{p}$, and if both $f$ and $g$ are contiguous and positive, then this inequality becomes an equality iff $f(x) = \norm{g}_{q}^{-\frac{1}{p-1}} g^{\frac{1}{p-1}}(x)$ almost everywhere.
\end{theorem}

Therefore, as long as $p > 1$, the Holder's inequality will always guarantee $W(v;\theta_W)$ converge to the closed-form formula $\norm{L}_{q}^{-\frac{1}{p-1}} L^{\frac{1}{p-1}}(v)$.

\subsection{Training Pipeline}\label{subsec:training-pipeline}
Our ARM can be easily attached to existing segmentation models.
As illustrated in Figure~\ref{fig:pipeline}, we first explicitly calculate the pixel-wise variance $v_i$ using the \textit{class-likelihood map}, and then normalize $v_i$ from $[0, \frac{C-1}{C^2}]$ to $[-1, 1]$.
Following that, we use ARM to map $v_i$ to $w_i$ and constrain the \textit{weight map} with $\norm{W}_p=1, p > 1$ for better convergence.
At last, we minimize the weighted loss sum of the \textit{weight map} and the \textit{loss map} as usual.
However, \textbf{by simply multiplying $-1$ to the learning rate of $\theta_W$, we achieve a kind of the adversarial training}.
While the segmentation model targets predicting more accurate class-likelihoods and producing a lower weighted loss;
our ARM tends to force the segmentation model to focus more on lower confidence regions and produce a higher weighted loss.
It is worth mentioning that Figure~\ref{fig:pipeline} shows the actual structure of our ARM, which only contains 53 learnable parameters.
This module is fairly lightweight and only brings negligible computational overhead.

    \section{Experiments} \label{sec:experiments}
    \subsection{Benchmarks and Evaluations}\label{subsec:benchmarks-and-evaluations}
We select two finely-labeled datasets and two coarsely-labeled ones to evaluate the effectiveness of our ARM.

\textbf{ADE20K.} The ADE20K dataset~\cite{zhou2017scene} is tasked for general scene parsing, which contains 150 semantic categories, including 35 stuff classes and 115 discrete objects.
This dataset is divided into 20,210/2,000/3,000 images for training, validation and testing.

\textbf{Cityscapes Fine.} The Cityscapes dataset~\cite{cordts2016cityscapes} targets for urban scene understanding, which contains 30 semantic categories but only 19 categories are used for evaluation.
It provides 5,000 finely-labeled images which are further divided into 2,975/500/1,525 images for training, validation and testing.
We name these images and the corresponding \textbf{fine} annotations as \textit{Cityscapes Fine}.

\textbf{Cityscapes Coarse.} The Cityscapes dataset also provides coarse annotations for the 5,000 images mentioned above.
We name these images and the corresponding \textbf{coarse} annotations as \textit{Cityscapes Coarse}.
For a fair comparison, the evaluation is also performed on Cityscapes Fine validation set.
Note that, the annotation type is the only difference between \textit{Cityscapes Fine} and \textit{Cityscapes Coarse}.

\textbf{Cityscapes Coarse Extension.} Except for the 5,000 images mentioned above, Cityscapes provides additional 20,000 coarsely-labeled images.
We name these coarsely-labeled images as \textit{Cityscapes Coarse Extension} and also perform the evaluation on Cityscapes Fine validation set.

Following previous works, Results are reported in three metrics, namely, overall accuracy (aAcc), class average accuracy (mAcc) and mean IoU (mIoU).

\subsection{Implementation Details}\label{subsec:implementation-details}
\textbf{Segmentation Network.}
We take DeepLabv3+~\cite{chen2018encoder} as the base model and only supervise the model with the cross-entropy loss.
Following~\cite{zhao2017pyramid}, we adopt the auxiliary loss~\cite{szegedy2015going} and assign its loss weight to $0.4$.
For Cityscapes, ResNet101~\cite{he2016deep} is selected as the backbone.
For ADE20K, the backbone is replaced with ResNeSt101~\cite{zhang2020resnest} for better performance.
All experiments are performed on 8 GTX2080Ti GPUs with mixed precision training.
The optimizer is SGD with a weight decay $0.0001$.
The learning rate policy is multi-step with warming up: given a base learning rate $\alpha_{0}$ and the total number of iterations $T$, the learning rate $\alpha$ will linearly increase from $\alpha_{0}/100$ to $\alpha_{0}$ in the first $0.05T$, and then decay to $\alpha_0/10$ at $0.6T$, and decay to $\alpha_0/100$ at $0.9T$.
We set $\alpha_0$ to $0.02$ for all experiments, while take $T$ as a dataset-related parameter.
With the batch size unified to 16, we set $T=40,000$ for Cityscapes Fine and Cityscapes Coarse (about 216 epochs), $T=160,000$ for ADE20K (about 127 epochs) and $T=200,000$ for Cityscapes Coarse Extension (about 160 epochs).

The data augmentation contains a series of randomly color-jitterring, randomly horizon-flipping, randomly resizing, and randomly cropping.
Only the image shape $(H, W)$ after randomly cropping is dataset-related, which is set to $(512, 1024)$ for Cityscapes, and $(512, 512)$ for ADE20K.
The code repository will be released later.

\textbf{Adversarial Reweighting Module.}
ARM is implemented by four 1$\times$1 convolution layers.
Its interlayer activation function is \textit{Swish}~\cite{ramachandran2017searching} and the final \textit{Sigmoid} is used to squeeze the sample weight to $[0, 1]$.
After grid-searching the optimal normalize type and the optimal input type in Table~\ref{tab:Search-hyper-parameters}, we set the input parameter of ARM to the normalized variance, and constrain ARM with $\norm{W}_2=1$.
The learning rate of ARM is always kept the same with the main segmentation model, except for an extra negative sign.
Considering the prediction confidence may be unreliable in the early training stage, we train ARM and the main model isolatedly in the first $0.05T$ and then jointly optimize them in the remaining iterations.

A dedicated training details is that the \textit{weight map} must be normalized twice.
Specifically, the \textit{weight map} is first divided by its \textit{L2 norm} to guarantee the convergence, and further divided by its \textit{L1 norm} to ensure the same magnitude of loss before and after reweighting.
Note that, both operations are indispensable.
Only dividing by L1 norm makes ARM fail to converge (see Table~\ref{tab:Search-hyper-parameters});
while only dividing by L2 norm makes the sum of sample weights greater than 1.
In other words, the learning rate is implicitly increased.

\subsection{Ablation Study}\label{subsec:ablation-study}
All experiments in this section are trained on the training set of each benchmark and evaluated on the corresponding validation set.
For simplicity, we only report the single-scale-testing results in this section.

\textbf{Linear ARM.}
Before grid-searching hyper-parameters for ARM, some fundamental experiments need to be performed first.
For this purpose, we introduce a simplified ARM as $W(v) = k*v + b$ with $k$ and $b$ taken as constants.
Since it is a linear mapping function, we name it as \textit{Linear ARM}.
The parameter $k$ plays a key role in the reweighting strategy.
A positive $k$ means assigning high-confidence samples a large weight and focusing on them, while a negative $k$ means suppressing these high-confidence samples.

\textbf{Detaching Gradient.}
An essential question is whether to detach the gradient of $p_{i,1:C}$ when calculating the variance $v_i$ and the sample weight $w_i$, more vividly, whether to tell the segmentation network that ARM is performing the reweighting task.
As shown in Table~\ref{tab:some-training-details-of-arm}, the mIoU of our base model (DeepLabv3plus + ResNet101) is 70.33.
However, after applying ARM to the segmentation model and calculating $w_i$ without detaching the gradient of $p_{i,1:C}$, the mIoU drops to 4.03.
Furthermore, once we detach the gradient, the mIoU increases to 72.13.

\begin{table}[b]
    \centering
    \caption{Some training details about our ARM.}
    \resizebox{\columnwidth}{!}{
        \begin{tabular}{lccrrr}
            \hline
            & \textbf{detach} & \textbf{aAcc}  & \textbf{mAcc}  & \textbf{mIoU}  \\ \hline
            Base model               &                 & 93.55          & 82.21          & 70.33          \\
            Linear ARM               &                 & 30.72          & 8.27           & 4.03           \\
            Linear ARM & $\checkmark$ & \textbf{93.92} & \textbf{83.30}
            & \textbf{72.13}
            \\ \hline
            Base model               &                 & 93.55          & 82.21          & 70.33          \\
            Linear ARM (suppressing) & $\checkmark$    & \textbf{93.92} & \textbf{83.30} & \textbf{72.13} \\
            Linear ARM (focusing)    & $\checkmark$    & 93.44          & 81.46          & 69.54          \\ \hline
        \end{tabular}
    }
    \label{tab:some-training-details-of-arm}
\end{table}

An intuition explanation is as follows:
If the segmentation model knows the existence of the ARM, then for $\theta_s$, the problem $\theta_s = \argmin_{\theta_s} \sum_{i} w_{i}l_{i}$ can be solved by two approaches.
The first one is minimizing $l_i$ by classifying a sample to the correct class, while the second one is minimizing $w_i$ by generating a higher variance $v_i$.
Obviously, directly generating a higher $v_i$ is much easier than classifying a sample to the correct class.
Therefore, when not detaching the gradient, our ARM unexpectedly provides a shortcut for the segmentation model to cheat.
But after detaching the gradient, our ARM obtains a significant performance improvement and increases the mIoU by 1.80.

\textbf{Suppressing \textit{vs} Focusing.}
Experiments are conducted with Linear ARM.
$W(v) = -0.5 * v + 0.5$ is used to suppress high-variance samples while $W(var) = 0.5 * var + 0.5$ is used to focus on them.
As shown in Table~\ref{tab:some-training-details-of-arm}, focusing on high-variance samples produces the lowest mIoU, and is even worse than our base model.

\textbf{The Convergence of Our ARM.}
To valid the theory provided in Sec.~\ref{subsec:convergence-of-ARM}, we perform experiments by constraining $\norm{W}_p=1$ with $p=1,2,3$, respectively.
As shown in Table~\ref{tab:Search-hyper-parameters}, the L1 norm does not converge, and the L2 norm produces the best mIoU.
This validates the conclusion in Sec.~\ref{subsec:convergence-of-ARM}: $q>1$ is critical to the convergence of our ARM.

\textbf{The Optimal Input Type.}
Except for the variance, other statistics about $p_{1:C}$ can also be used to indicate the confidence, like $std$, $\log var$ and $entropy=\sum_{i} -p_i\log p_i$.
As shown in Table~\ref{tab:Search-hyper-parameters}, the performance gap between $var$, $std$ and $\log var$ is very small.
However, setting the input type to $entropy$ makes ARM not converge, which may be caused by its numerical instability.

\begin{table}
    \centering
    \caption{Search hyper-parameters for our ARM.}
    \resizebox{\columnwidth}{!}{
        \begin{tabular}{ccccc}
            \hline
            \textbf{Hyper-parameter}        & \textbf{subtype} & \textbf{aAcc}  & \textbf{mAcc}  & \textbf{mIoU}  \\ \hline
            \multirow{4}{*}{Normalize type} & Base model       & 93.55          & 82.21          & 70.33          \\
            & \textit{L1 norm} & NaN            & NaN            & NaN            \\
            & \textit{L2 norm} & \textbf{93.88} & \textbf{83.83} & \textbf{71.94} \\
            & \textit{L3 norm} & 93.84          & 83.39          & 71.10          \\ \hline
            \multirow{5}{*}{Input type}     & Base model       & 93.55          & 82.21          & 70.33          \\
            & $var$            & \textbf{93.88} & \textbf{83.83} & \textbf{71.94} \\
            & $std$            & 93.88          & 83.46          & 71.83          \\
            & $\log var$       & 93.84          & 83.27          & 71.93          \\
            & $entropy$        & NaN            & NaN            & NaN            \\ \hline
        \end{tabular}
    }
    \label{tab:Search-hyper-parameters}
\end{table}

\textbf{Applying ARM to Different Backbones.}
We report the model performance with different backbones, namely, ResNet50 and ResNet101.
As shown in Table~\ref{tab:applying-to-different-models-and-backbones}, ARM brings consistent performance improvement, and the mIoU of ResNet50 enhanced by our ARM outperforms the vanilla ResNet101 on both Cityscapes Coarse and Cityscapes Fine.

\textbf{Applying ARM to Different Heads.}
To prove our ARM is model-independent, we attach it to different segmentation heads, namely, DeepLabv3+ and OCR~\cite{yuan2019object}.
As shown in Table~\ref{tab:applying-to-different-models-and-backbones}, our ARM brings consistent performance improvement on different models and different annotation types.
Though adopting the same hyper-parameters for all experiments affects the performance of OCR, the relative improvement brought by ARM is still acceptable.

\begin{table}
    \centering
    \caption{Applying ARM to different backbones and heads.}\label{tab:applying-to-different-models-and-backbones}
    \resizebox{\columnwidth}{!}{
        \begin{tabular}{cllccc}
            \hline
            \textbf{Dataset} & \textbf{Backbone type} & \textbf{Model} & \textbf{aAcc} & \textbf{mAcc}
            & \textbf{mIoU}
            \\ \hline
            \multirow{4}{*}{Cityscapes Coarse} & \multirow{2}{*}{ResNet50} & Base model & 93.56 & 82.33
            & 69.53
            \\
            &                                            & ARM            & \textbf{93.75} & \textbf{82.46} & \textbf{70.98} \\ \cline{2-6}
            & \multirow{2}{*}{ResNet101}                 & Base model     & 93.55          & 82.21          & 70.33          \\
            &                                            & ARM            & \textbf{93.88} & \textbf{83.83} & \textbf{71.94} \\ \hline
            \multirow{4}{*}{Cityscapes Fine} & \multirow{2}{*}{ResNet50} & Base model & 96.34 & 90.08
            & 79.32
            \\
            &                                            & ARM            & \textbf{96.52} & \textbf{90.39} & \textbf{80.86} \\ \cline{2-6}
            & \multirow{2}{*}{ResNet101}                 & Base model     & 96.49          & 90.47          & 80.76          \\
            &                                            & ARM            & \textbf{96.62} & \textbf{90.97} & \textbf{81.73} \\ \hline
            \textbf{Dataset} & \textbf{Head type}                         & \textbf{Model} & \textbf{aAcc}  & \textbf{mAcc}  & \textbf{mIoU}  \\ \hline
            \multirow{4}{*}{Cityscapes Coarse} & \multirow{2}{*}{DeepLabv3+~\cite{chen2018encoder}}
            & Base model
            & 93.55
            & 82.21
            & 70.33
            \\
            &                                            & ARM            & \textbf{93.88} & \textbf{83.83} & \textbf{71.94} \\ \cline{2-6}
            & \multirow{2}{*}{OCR~\cite{yuan2019object}} & Base model     & 93.65          & 82.52          & 70.51          \\
            &                                            & ARM            & \textbf{93.77} & \textbf{83.31} & \textbf{71.79} \\ \hline
            \multirow{4}{*}{Cityscapes Fine} & \multirow{2}{*}{DeepLabv3+~\cite{chen2018encoder}}
            & Base model
            & 96.49
            & 90.47
            & 80.76
            \\
            &                                            & ARM            & \textbf{96.62} & \textbf{90.97} & \textbf{81.73} \\ \cline{2-6}
            & \multirow{2}{*}{OCR~\cite{yuan2019object}} & Base model     & 96.39          & 90.17          & 79.92          \\
            &                                            & ARM            & \textbf{96.46} & \textbf{90.94} & \textbf{80.60} \\ \hline
        \end{tabular}
    }
\end{table}

\textbf{Applying ARM to Different Loss Functions.}
We perform experiments with the cross-entropy loss (CE), the robust loss function Generalized CE loss (GCE)~\cite{zhang2018generalized}, and the KL loss mentioned above.
For GCE, we grid-search its hyper-parameter $q$ and find the optimal $q$ is $0.1$.
After that, we combine GCE with ARM by setting ARM's input parameter to $var$.
For KL, we combine it with ARM by setting ARM's input parameter to the $s$ term in KL loss.
From Table~\ref{tab:applying-to-different-loss-functions}, we can observe that the base model supervised by different losses show a clear performance gap.
GCE performs the best, and ``KL is worse than CE'' validates the analysis in Sec.\ref{subsec:dive-into-kl-loss}.
By applying our ARM to them, all of these three loss functions obtain significant performance improvements.
This strongly proves the robustness and effectiveness of our ARM.

\begin{table}
    \centering
    \caption{Applying ARM to different loss functions.}
    \resizebox{\columnwidth}{!}{
        \begin{tabular}{lcccl}
            \hline
            \textbf{Loss type} & \textbf{use ARM} & \textbf{aAcc}  & \textbf{mAcc}  & \textbf{mIoU}         \\ \hline
            KL                 &                  & 93.44          & 81.15          & 69.06                 \\
            CE                 &                  & 93.55          & 82.21          & 70.33                 \\
            GCE                &                  & 93.69          & 82.53          & 71.23                 \\ \hline
            KL                 & $\checkmark$     & \textbf{93.83} & \textbf{83.01} & \textbf{71.86(+2.80)} \\
            CE                 & $\checkmark$     & \textbf{93.88} & \textbf{83.83} & \textbf{71.94(+1.61)} \\
            GCE                & $\checkmark$     & \textbf{93.85} & \textbf{83.28} & \textbf{72.08(+0.85)} \\ \hline
        \end{tabular}
    }
    \label{tab:applying-to-different-loss-functions}
\end{table}

\textbf{Comparison with OHEM}\label{subsec:comparison-with-ohem}
We perform experiments to prove our ARM is better than OHEM.
The hyper-parameter of OHEM is adopted from~\cite{yuan2019object}.
As shown in Table~\ref{tab:comparison-with-ohem}, OHEM only brings performance improvements on Cityscapes Fine and ADE20K, but does not on Cityscapes Coarse.
Though ARM performs slightly worse than OHEM on ADE20K, it brings consistent performance improvements on all datasets.
This shows that our ARM is more robust and more effective than OHEM.

\begin{table}
    \centering
    \caption{Comparison with OHEM on the validation set of each dataset with single-scale-testing.}
    \resizebox{\columnwidth}{!}{
        \begin{tabular}{lllcl}
            \hline
            \textbf{Dataset}                   & \textbf{Method} & \textbf{aAcc} & \textbf{mAcc} & \textbf{mIoU} \\ \hline
            \multirow{3}{*}{Cityscapes Fine}   & Base model      & 96.49         & 90.47         & 80.76         \\
            & OHEM            & 96.52         & 90.74         & 81.23(+0.47)  \\
            & ARM             & 96.62         & 90.97         & 81.73(+0.97)  \\ \hline
            \multirow{3}{*}{ADE20K}            & Base model      & 80.88         & 63.55         & 45.91         \\
            & OHEM            & 81.25         & 65.19         & 46.78(+0.87)  \\
            & ARM             & 81.07         & 64.22         & 46.51(+0.60)  \\ \hline \hline
            \multirow{3}{*}{Cityscapes Coarse} & Base model      & 93.55         & 82.21         & 70.33         \\
            & OHEM            & 93.64         & 82.23         & 70.31(-0.02)  \\
            & ARM             & 93.88         & 83.83         & 71.94(+1.61)  \\ \hline
        \end{tabular}
    }
    \label{tab:comparison-with-ohem}
\end{table}

\subsection{Comparison with SOTA}\label{subsec:comparison-with-sota}

\begin{table}
    \centering
    \caption{
        Comparison with SOTA. We use $\dag$ and $*$ to mark the results using the Mapillary dataset and the results using the ResNeSt101 backbone.
        All methods adopt the same TTA.
    }
    \resizebox{\columnwidth}{!}{
        \begin{tabular}{lclll}
            \hline
            \multirow{2}{*}{\textbf{Method}} & \multirow{2}{*}{\textbf{Backbone}} & \multicolumn{2}{l}{\textbf{\ \ \ \ \ \ \ \ \ \ Cityscapes}}
            & \multirow{2}{*}{\textbf{ADE20K}}
            \\
            &              & Fine                  & Coarse                &                           \\
            \hline
            CPNet~\cite{yu2020context}            & ResNet101    & 81.3                  & \ \ \ -               & 46.27                     \\
            BFP~\cite{ding2019boundary}           & ResNet101    & 81.4                  & \ \ \ -               & \ \ \ -                   \\
            OCNet~\cite{yuan2018ocnet}            & ResNet101    & 81.7                  & \ \ \ -               & 45.45                     \\
            ACFNet~\cite{zhang2019acfnet}         & ResNet101    & 81.8                  & \ \ \ -               & \ \ \ -                   \\
            GALD~\cite{li2019global}              & ResNet101    & 81.8                  & \ \ \ -               & \ \ \ -                   \\
            OCR~\cite{yuan2019object}             & ResNet101    & 81.8                  & \ \ \ -               & 45.28                     \\
            DGCNet~\cite{zhang2019dual}           & ResNet101    & 82.0                  & \ \ \ -               & \ \ \ -                   \\
            ACNet~\cite{fu2019adaptive}           & ResNet101    & 82.3                  & \ \ \ -               & 45.90                     \\
            OCR~\cite{yuan2019object}             & HRNetV2-W48  & 82.4                  & \ \ \ -               & 45.66                     \\
            DPC~\cite{chen2018searching}          & Xception-71  & 82.7                  & \ \ \ -               & \ \ \ -                   \\
            GSCNN$^\dag$~\cite{takikawa2019gated} & WideResNet38 & 82.8                  & \ \ \ -               & \ \ \ -                   \\
            Decouple~\cite{li2020improving}       & ResNet101    & 82.8                  & \ \ \ -               & \ \ \ -                   \\
            DRANet~\cite{fu2020scene}             & ResNet101    & \textbf{82.9}         & \ \ \ -               & 46.18                     \\
            ResNeSt~\cite{zhang2020resnest}       & ResNeSt101   & \ \ \ -               & \ \ \ -               & 46.91$^*$                 \\
            \hline
            Base model                            & ResNet101    & 81.39                 & 71.28                 & 46.46$^*$                 \\
            OHEM~\cite{shrivastava2016training}   & ResNet101    & 81.81(+0.42)          & 71.18(-0.10)          & 47.30(+0.84)$^*$          \\
            Linear ARM                            & ResNet101    & \textbf{82.05(+0.66)} & 71.63(+0.35)          & \textbf{47.50(+1.04)}$^*$ \\
            ARM                                   & ResNet101    & 81.87(+0.48)          & \textbf{71.96(+0.68)} & 47.19(+0.63)$^*$          \\ \hline
        \end{tabular}
    }
    \label{tab:comparison-with-sota}
\end{table}

We adopt the same test time augmentation (TTA) as~\cite{yuan2019object}, which contains multi-scale testing and horizon flipping.
Results of Cityscapes Fine and Cityscapes Coarse are trained on \texttt{train+val} and evaluated on \texttt{test} online.
Results of ADE20K are trained on \texttt{train} and evaluated on \texttt{val}.

Specifically, for Cityscapes Fine and Cityscapes Coarse, we take DeepLabv3plus + ResNet101 as the base model and train it on \texttt{train+val} with $T=40,000$.
Note that, Cityscapes Fine and Cityscapes share the same training images, but are labeled with different annotation type.
For ADE20K, we set $T=160,000$ and replace the backbone with ResNeSt101~\cite{zhang2020resnest} for better performance.

As shown in Table~\ref{tab:comparison-with-sota}, OHEM only performs well on finely-labeled datasets, while our ARM and Linear ARM bring consistent improvements on all datasets and obtain a performance comparable to SOTAs.
This observation is consistent with the analysis in the introduction and strongly proves that the confidence-based reweighting strategies are more robust and effective than loss-based ones.
Moreover, Linear ARM seems to perform better on the finely-labeled Cityscapes Fine and ADE20K, while ARM does better on Cityscapes Coarse.
We may conclude that a linear reweighting function is enough for noise-free datasets, while noisy datasets need a more complicated reweighting function.

Note that, with the help of ResNeSt101, Linear ARM achieves a mIoU of 47.50 on the ADE20K validation set, which outperforms all other methods.

\subsection{Discussion}\label{subsec:discussion}
\textbf{Linear ARM \textit{vs} ARM.}
Both Linear ARM and ARM are just an implementation of $W(v; \theta_W)$.
Linear ARM is designed with only two fixed constants $k$ and $b$ and is not learnable;
while ARM is designed with 53 learnable parameters and is optimized by a negative learning rate.

Though ARM may be inferior to Linear ARM sometimes, we still think it is interesting to automatically learning an acceptable $W(v; \theta_W)$ with only a negative learning rate.
Besides, both OHEM and Linear ARM need to perform \textbf{multiple} rounds of experiments for manual tuning the hyper-parameters.
However, ARM can automatically obtain an acceptable mapping function by only \textbf{one} round of experiments.
Furthermore, the existing ARM only introduces a negligible amount of calculation, and there is still great potential for mining.

\textbf{Coarse Semantic Segmentation.}
Looking back now, our original intention is to tailor-make a reweighting strategy for coarse annotations and narrow the performance gap between two annotation types.
However, as illustrated in Table~\ref{tab:comparison-with-sota}, our ARM actually brings similar performance improvement for both annotation types.
The mIoU gap between these two annotation types remains about 10 points, which is a little disappointing.

So does there still exist potentials for coarse semantic segmentation?
We answer yes for two reason.
First, the coarse annotation is easy to obtain.
While the fine annotation requires nearly 1.5 hours to label, the coarse one takes just 7 minutes, which is $1/13$ of the former.
The mIoU of 82.05 is achieved by 2975+500 finely-labeled images with a labeling time of 5,175 hours.
As long as we can use coarse annotations achieve this mIoU within 5,175 hours, the task \textit{coarse semantic segmentation} is still meaningful.
In our appendix, we increase the mIoU from 71.96 to 73.90 with 19,998 images in \textit{Cityscapes Coarse Extension} (2,333 hours to label) and narrow the mIoU gap from 10.09 to 8.15.
Second, the reweighting strategy is not the only way to solve the noise problem.
Identifying and correcting mislabeled pixels may further boost the performance of coarse annotations.
To conclude, we think coarse semantic segmentation is an interesting task and worthy of more in-depth analysis.

    \section{Conclusion}\label{sec:conclusion}
    In this paper, we take coarsely-labeled images as a noisy semantic segmentation dataset and suppress the label noise with a reweighting strategy.
After pointing out the dilemma faced by loss-based reweighting strategies, we propose a confidence-based one and generalize it to an adversarial reweighting module named ARM.
Extensive experiments on different datasets and different annotation types prove the effectiveness and robustness of our ARM.

    \bibliographystyle{ieee_fullname}
    \bibliography{ms.bbl}

\begin{thebibliography}{10}\itemsep=-1pt

\bibitem{cai2020learning}
Qi Cai, Yingwei Pan, Yu Wang, Jingen Liu, Ting Yao, and Tao Mei.
\newblock Learning a unified sample weighting network for object detection.
\newblock In {\em Proceedings of the IEEE/CVF Conference on Computer Vision and
  Pattern Recognition}, pages 14173--14182, 2020.

\bibitem{cao2020prime}
Yuhang Cao, Kai Chen, Chen~Change Loy, and Dahua Lin.
\newblock Prime sample attention in object detection.
\newblock In {\em CVPR}, pages 11583--11591, 2020.

\bibitem{chang2017active}
Haw-Shiuan Chang, Erik Learned-Miller, and Andrew McCallum.
\newblock Active bias training more accurate neural networks by emphasizing
  high variance samples.
\newblock In {\em Advances in Neural Information Processing Systems}, pages
  1002--1012, 2017.

\bibitem{chen2018searching}
Liang-Chieh Chen, Maxwell~D Collins, Yukun Zhu, George Papandreou, Barret Zoph,
  Florian Schroff, Hartwig Adam, and Jonathon Shlens.
\newblock Searching for efficient multi-scale architectures for dense image
  prediction.
\newblock {\em NeurIPS}, 2018.

\bibitem{chen2018encoder}
Liang-Chieh Chen, Yukun Zhu, George Papandreou, Florian Schroff, and Hartwig
  Adam.
\newblock Encoder-decoder with atrous separable convolution for semantic image
  segmentation.
\newblock In {\em Proceedings of the European conference on computer vision
  (ECCV)}, pages 801--818, 2018.

\bibitem{cordts2016cityscapes}
Marius Cordts, Mohamed Omran, Sebastian Ramos, Timo Rehfeld, Markus Enzweiler,
  Rodrigo Benenson, Uwe Franke, Stefan Roth, and Bernt Schiele.
\newblock The cityscapes dataset for semantic urban scene understanding.
\newblock In {\em Proceedings of the IEEE conference on computer vision and
  pattern recognition}, pages 3213--3223, 2016.

\bibitem{ding2019boundary}
Henghui Ding, Xudong Jiang, Ai~Qun Liu, Nadia~Magnenat Thalmann, and Gang Wang.
\newblock Boundary-aware feature propagation for scene segmentation.
\newblock In {\em Proceedings of the IEEE/CVF International Conference on
  Computer Vision}, pages 6819--6829, 2019.

\bibitem{fu2020scene}
Jun Fu, Jing Liu, Jie Jiang, Yong Li, Yongjun Bao, and Hanqing Lu.
\newblock Scene segmentation with dual relation-aware attention network.
\newblock {\em IEEE Transactions on Neural Networks and Learning Systems},
  2020.

\bibitem{fu2019adaptive}
Jun Fu, Jing Liu, Yuhang Wang, Yong Li, Yongjun Bao, Jinhui Tang, and Hanqing
  Lu.
\newblock Adaptive context network for scene parsing.
\newblock In {\em Proceedings of the IEEE/CVF International Conference on
  Computer Vision}, pages 6748--6757, 2019.

\bibitem{holder1889ueber}
Otto H~{\"o} lder.
\newblock Ueber einen mittelwertsatz.
\newblock {\em Nachrichten von der K {\"o} nig. Gesellschaft der Wissenschaften
  und der Georg-Augusts-Universit {\"a} t zu G {\"o} ttigen}, pages 38--47,
  1889.

\bibitem{han2018masking}
Bo Han, Jiangchao Yao, Gang Niu, Mingyuan Zhou, Ivor Tsang, Ya Zhang, and
  Masashi Sugiyama.
\newblock Masking: A new perspective of noisy supervision.
\newblock In {\em Advances in Neural Information Processing Systems}, pages
  5836--5846, 2018.

\bibitem{he2016deep}
Kaiming He, Xiangyu Zhang, Shaoqing Ren, and Jian Sun.
\newblock Deep residual learning for image recognition.
\newblock In {\em Proceedings of the IEEE conference on computer vision and
  pattern recognition}, pages 770--778, 2016.

\bibitem{he2019bounding}
Yihui He, Chenchen Zhu, Jianren Wang, Marios Savvides, and Xiangyu Zhang.
\newblock Bounding box regression with uncertainty for accurate object
  detection.
\newblock In {\em Proceedings of the IEEE Conference on Computer Vision and
  Pattern Recognition}, pages 2888--2897, 2019.

\bibitem{jiang2018mentornet}
Lu Jiang, Zhengyuan Zhou, Thomas Leung, Li-Jia Li, and Li Fei-Fei.
\newblock Mentornet: Learning data-driven curriculum for very deep neural
  networks on corrupted labels.
\newblock In {\em ICML}, pages 2304--2313, 2018.

\bibitem{kendall2017uncertainties}
Alex Kendall and Yarin Gal.
\newblock What uncertainties do we need in bayesian deep learning for computer
  vision?
\newblock In {\em Advances in neural information processing systems}, pages
  5574--5584, 2017.

\bibitem{kendall2018multi}
Alex Kendall, Yarin Gal, and Roberto Cipolla.
\newblock Multi-task learning using uncertainty to weigh losses for scene
  geometry and semantics.
\newblock In {\em CVPR}, pages 7482--7491, 2018.

\bibitem{khoreva2017simple}
Anna Khoreva, Rodrigo Benenson, Jan Hosang, Matthias Hein, and Bernt Schiele.
\newblock Simple does it: Weakly supervised instance and semantic segmentation.
\newblock In {\em Proceedings of the IEEE conference on computer vision and
  pattern recognition}, pages 876--885, 2017.

\bibitem{kulharia2020box2seg}
Viveka Kulharia, Siddhartha Chandra, Amit Agrawal, Philip Torr, and Ambrish
  Tyagi.
\newblock Box2seg: Attention weighted loss and discriminative feature learning
  for weakly supervised segmentation.
\newblock In {\em European Conference on Computer Vision}, pages 290--308.
  Springer, 2020.

\bibitem{li2020improving}
Xiangtai Li, Xia Li, Li Zhang, Guangliang Cheng, Jianping Shi, Zhouchen Lin,
  Shaohua Tan, and Yunhai Tong.
\newblock Improving semantic segmentation via decoupled body and edge
  supervision.
\newblock {\em arXiv preprint arXiv:2007.10035}, 2020.

\bibitem{li2019global}
Xiangtai Li, Li Zhang, Ansheng You, Maoke Yang, Kuiyuan Yang, and Yunhai Tong.
\newblock Global aggregation then local distribution in fully convolutional
  networks.
\newblock {\em BMVC}, 2019.

\bibitem{li2017learning}
Yuncheng Li, Jianchao Yang, Yale Song, Liangliang Cao, Jiebo Luo, and Li-Jia
  Li.
\newblock Learning from noisy labels with distillation.
\newblock In {\em ICCV}, pages 1910--1918, 2017.

\bibitem{lin2017focal}
Tsung-Yi Lin, Priya Goyal, Ross Girshick, Kaiming He, and Piotr Doll~{\'a} r.
\newblock Focal loss for dense object detection.
\newblock In {\em ICCV}, pages 2980--2988, 2017.

\bibitem{luo2018coarse}
Yadan Luo, Ziwei Wang, Zi Huang, Yang Yang, and Cong Zhao.
\newblock Coarse-to-fine annotation enrichment for semantic segmentation
  learning.
\newblock In {\em Proceedings of the 27th ACM International Conference on
  Information and Knowledge Management}, pages 237--246, 2018.

\bibitem{ma2020normalized}
Xingjun Ma, Hanxun Huang, Yisen Wang, Simone Romano, Sarah Erfani, and James
  Bailey.
\newblock Normalized loss functions for deep learning with noisy labels.
\newblock {\em ICML}, 2020.

\bibitem{ramachandran2017searching}
Prajit Ramachandran, Barret Zoph, and Quoc~V Le.
\newblock Searching for activation functions.
\newblock {\em arXiv preprint arXiv:1710.05941}, 2017.

\bibitem{ren2018learning}
Mengye Ren, Wenyuan Zeng, Bin Yang, and Raquel Urtasun.
\newblock Learning to reweight examples for robust deep learning.
\newblock {\em ICML}, 2018.

\bibitem{shrivastava2016training}
Abhinav Shrivastava, Abhinav Gupta, and Ross Girshick.
\newblock Training region-based object detectors with online hard example
  mining.
\newblock In {\em CVPR}, pages 761--769, 2016.

\bibitem{shu2019meta}
Jun Shu, Qi Xie, Lixuan Yi, Qian Zhao, Sanping Zhou, Zongben Xu, and Deyu Meng.
\newblock Meta-weight-net: Learning an explicit mapping for sample weighting.
\newblock In {\em Advances in Neural Information Processing Systems}, pages
  1919--1930, 2019.

\bibitem{song2020learning}
Hwanjun Song, Minseok Kim, Dongmin Park, and Jae-Gil Lee.
\newblock Learning from noisy labels with deep neural networks: A survey.
\newblock {\em arXiv preprint arXiv:2007.08199}, 2020.

\bibitem{szegedy2015going}
Christian Szegedy, Wei Liu, Yangqing Jia, Pierre Sermanet, Scott Reed, Dragomir
  Anguelov, Dumitru Erhan, Vincent Vanhoucke, and Andrew Rabinovich.
\newblock Going deeper with convolutions.
\newblock In {\em Proceedings of the IEEE conference on computer vision and
  pattern recognition}, pages 1--9, 2015.

\bibitem{takikawa2019gated}
Towaki Takikawa, David Acuna, Varun Jampani, and Sanja Fidler.
\newblock Gated-scnn: Gated shape cnns for semantic segmentation.
\newblock In {\em Proceedings of the IEEE/CVF International Conference on
  Computer Vision}, pages 5229--5238, 2019.

\bibitem{tian2020boxinst}
Zhi Tian, Chunhua Shen, Xinlong Wang, and Hao Chen.
\newblock Boxinst: High-performance instance segmentation with box annotations.
\newblock {\em arXiv preprint arXiv:2012.02310}, 2020.

\bibitem{wang2017robust}
Yixin Wang, Alp Kucukelbir, and David~M Blei.
\newblock Robust probabilistic modeling with bayesian data reweighting.
\newblock In {\em ICML}, pages 3646--3655, 2017.

\bibitem{yu2020context}
Changqian Yu, Jingbo Wang, Changxin Gao, Gang Yu, Chunhua Shen, and Nong Sang.
\newblock Context prior for scene segmentation.
\newblock In {\em CVPR}, 2020.

\bibitem{yuan2019object}
Yuhui Yuan, Xilin Chen, and Jingdong Wang.
\newblock Object-contextual representations for semantic segmentation.
\newblock In {\em ECCV}, 2020.

\bibitem{yuan2018ocnet}
Yuhui Yuan and Jingdong Wang.
\newblock Ocnet: Object context network for scene parsing.
\newblock {\em arXiv preprint arXiv:1809.00916}, 2018.

\bibitem{zhang2019acfnet}
Fan Zhang, Yanqin Chen, Zhihang Li, Zhibin Hong, Jingtuo Liu, Feifei Ma, Junyu
  Han, and Errui Ding.
\newblock Acfnet: Attentional class feature network for semantic segmentation.
\newblock In {\em Proceedings of the IEEE/CVF International Conference on
  Computer Vision}, pages 6798--6807, 2019.

\bibitem{zhang2020resnest}
Hang Zhang, Chongruo Wu, Zhongyue Zhang, Yi Zhu, Zhi Zhang, Haibin Lin, Yue
  Sun, Tong He, Jonas Mueller, R Manmatha, et~al.
\newblock Resnest: Split-attention networks.
\newblock {\em arXiv preprint arXiv:2004.08955}, 2020.

\bibitem{zhang2019dual}
Li Zhang, Xiangtai Li, Anurag Arnab, Kuiyuan Yang, Yunhai Tong, and Philip~HS
  Torr.
\newblock Dual graph convolutional network for semantic segmentation.
\newblock {\em arXiv preprint arXiv:1909.06121}, 2019.

\bibitem{zhang2018generalized}
Zhilu Zhang and Mert Sabuncu.
\newblock Generalized cross entropy loss for training deep neural networks with
  noisy labels.
\newblock In {\em Advances in neural information processing systems}, pages
  8778--8788, 2018.

\bibitem{zhao2017pyramid}
Hengshuang Zhao, Jianping Shi, Xiaojuan Qi, Xiaogang Wang, and Jiaya Jia.
\newblock Pyramid scene parsing network.
\newblock In {\em Proceedings of the IEEE conference on computer vision and
  pattern recognition}, pages 2881--2890, 2017.

\bibitem{zheng2020rectifying}
Zhedong Zheng and Yi Yang.
\newblock Rectifying pseudo label learning via uncertainty estimation for
  domain adaptive semantic segmentation.
\newblock {\em arXiv preprint arXiv:2003.03773}, 2020.

\bibitem{zhou2017scene}
Bolei Zhou, Hang Zhao, Xavier Puig, Sanja Fidler, Adela Barriuso, and Antonio
  Torralba.
\newblock Scene parsing through ade20k dataset.
\newblock In {\em Proceedings of the IEEE conference on computer vision and
  pattern recognition}, pages 633--641, 2017.

\bibitem{zlateski2018importance}
Aleksandar Zlateski, Ronnachai Jaroensri, Prafull Sharma, and Fr~{\'e}~do
  Durand.
\newblock On the importance of label quality for semantic segmentation.
\newblock In {\em Proceedings of the IEEE Conference on Computer Vision and
  Pattern Recognition}, pages 1479--1487, 2018.

\end{thebibliography}

\end{document}